\documentclass{article}
\usepackage[font=footnotesize,labelfont=bf]{caption}
\usepackage[utf8]{inputenc}
\usepackage[T1]{fontenc}
\usepackage{amsmath,amssymb,amsfonts,bm}
\usepackage{hyperref}
\usepackage{listings}
\usepackage{xcolor}
\usepackage{graphicx}
\usepackage{wrapfig}
\usepackage[subrefformat=parens,labelformat=parens]{subcaption}
\usepackage{cite}
\usepackage[ruled,lined,noend]{algorithm2e}
\usepackage{multirow}
\usepackage{array}
\usepackage{multicol}
\usepackage{longtable}
\usepackage{booktabs}
\usepackage{orcidlink}
\usepackage{amsthm}

\DeclareMathOperator*{\argmax}{argmax}

\hypersetup{
    colorlinks=true,
    linkcolor=blue,
    filecolor=magenta,
    urlcolor=cyan,
    citecolor=purple,
}

\SetKw{Break}{break}
\SetKw{Continue}{continue}
\SetKw{Let}{let}
\SetKw{In}{in}
\SetKw{Assert}{assert}

\SetKwBlock{Loop}{loop}{end}

\SetKwComment{Comment}{$\triangleright$\ }{}
\SetCommentSty{text}

\theoremstyle{plain}
\newtheorem{definition}{Definition}
\theoremstyle{definition}
\newtheorem{example}{Example}

\newcommand{\Hide}[1]{}

\newcommand{\Todo}[1]{{\em\color{red} #1}}

\newcommand*{\QED}{\hfill $\square$}

\newcommand*{\Relu}{ReLU}

\newcommand*{\ImpSpace}[1]{\mathbb{F}_{#1}}

\newcommand*{\Itp}{\mathit{Itp}}

\newcommand*{\ItpType}[1]{\Itp{}_\mathit{#1}}
\newcommand*{\ItpM}{\ItpType{M}}
\newcommand*{\ItpF}{\ItpType{F}}
\newcommand*{\ItpDF}{\ItpType{DF}}
\newcommand*{\ItpFactor}{\ItpType{f}}

\newcommand*{\Ucore}{\mathit{Ucore}}

\newcommand*{\Sample}{\mathbf{s}}
\newcommand*{\SampleExpl}{\varphi_\Sample{}}

\newcommand*{\LRA}{\textsf{QF\_LRA}}

\newcommand{\Opensmt}{\textsc{OpenSMT2}}
\newcommand{\Marabou}{\textsc{Marabou}}
\newcommand{\verix}{\textsc{VeriX}}
\newcommand{\spexplain}{\textsc{S\textsubscript{p}EX\textsubscript{pl}AI\textsubscript{n}}}

\newcommand*{\inst}[1]{\footnotemark[#1]\ \:}
\makeatletter
\renewcommand*\@fnsymbol[1]{\the#1}
\makeatother

\title{Space Explanations of Neural Network Classification}
\author{%
Faezeh Labbaf\inst{1}\orcidlink{0000-0002-8812-6702}, Tomáš Kolárik\inst{1}\orcidlink{0000-0002-7207-5197}, Martin Blicha\inst{1}\inst{4}\orcidlink{0000-0001-8140-4098},\\Grigory Fedyukovich\inst{1}\inst{3}\orcidlink{0000-0003-1727-4043}, Michael Wand\inst{2}\orcidlink{0000-0003-0966-7824}, Natasha Sharygina\inst{1}\orcidlink{0000-0002-8872-4913}%
\\[2em]
\small
\inst{1} University of Lugano (USI), Lugano, Switzerland
\\
\small
\inst{2} SUPSI, IDSIA, Lugano, Switzerland
\\
\small
\inst{3} Florida State University, United States of America
\\
\small
\inst{4} Ethereum Foundation
}

\date{}

\begin{document}

\maketitle
\setcounter{footnote}{0}

\begin{abstract}
We present a novel logic-based concept called \emph{Space Explanations} for classifying neural networks that gives provable guarantees of the behavior of the network in continuous areas of the input feature space.
To automatically generate space explanations, we leverage a range of flexible Craig interpolation algorithms and unsatisfiable core generation.
Based on real-life case studies, ranging from small to medium to large size, we demonstrate that the generated explanations are more meaningful than those computed by state-of-the-art.
\end{abstract}

\section{Introduction}

Explainability of decision-making AI systems (XAI),
and specifically neural networks (NNs), is a key requirement for deploying AI in sensitive areas~\cite{EUGroup_EthicsGuidelines}.
A~recent trend in explaining NNs is based on formal methods and logic,
providing explanations for the decisions
of machine learning systems~\cite{Shih_IJCAI18_SymbolicApproach,Seshia_TowardsVerifiedAI,Marques-Silva2023,Wu_2023,Ignatiev:24,LaMalfa_IJCAI21_GuaranteedOptimalRobustExplanations}
accompanied by provable guarantees regarding their correctness. 
Yet,
rigorous exploration of the \emph{continuous} feature space requires
to estimate \emph{decision boundaries} with complex shapes.
This, however, remains a~challenge
because existing explanations~\cite{Shih_IJCAI18_SymbolicApproach,Seshia_TowardsVerifiedAI,Marques-Silva2023,Wu_2023,Ignatiev:24,LaMalfa_IJCAI21_GuaranteedOptimalRobustExplanations}
constrain only individual features and hence
fail capturing \emph{relationships} among the features
that are essential to understand the reasons
behind the multi-parametrized classification process.

\Hide{
\begin{wrapfigure}{R}{0.45\textwidth}
    \centering
    \includegraphics[width=.45\textwidth]{figures/intro_relation.png}
    \caption{Explanation of misclassification}
    \label{fig:intro:bnd}
\end{wrapfigure}
}

\begin{figure}[t!]
    \centering
        \centering
        \includegraphics[trim={0cm 0cm 0cm 0cm},clip,width=.45\textwidth]{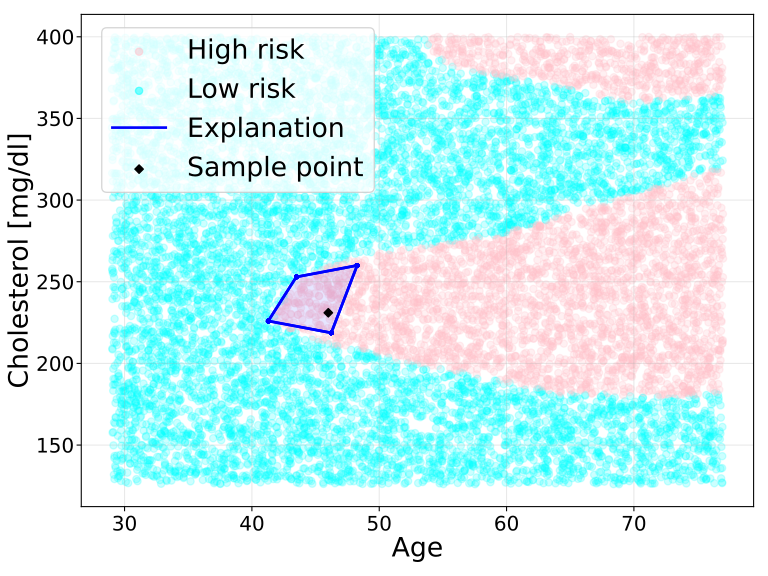}
    \caption{
    Example of close approximations of non-trivial decision boundaries
    using meaningful explanations of the classification: a~NN classifier of heart attack risk~\cite{heart_disease_45}.}
    \label{fig:bnd:real}
\end{figure}

We address the need to provide interpretations of NN systems
that are as meaningful as possible
using a~novel concept of \emph{Space Explanations},
delivered by a~flexible symbolic reasoning framework
where Craig interpolation \cite{Cra57} is at the heart of the machinery.
When starting from a sample point, the explanation computation is quick, and yields generality and soundness of approximation.
Interpolation extracts information from the proof of unsatisfiability which captures a reason why the given input cannot change classification.
We showcase explanations for trained neural networks
substantially more expressive and meaningful
compared to prior work in formal XAI.
For example,
the classification of the heart attack risk
is determined (among others) based on a~non-trivial relation
of age and cholesterol level of a patient, as illustrated in Figure~\ref{fig:bnd:real}.
Our prototype implementation of the framework,
\spexplain{},
can automatically compute an explanation that
closely approximates the shape of the decision boundary.
Furthermore,
such non-restricted explanations often cover
many sample points at once.
\Hide{
Moreover,
multiple explanations can approximate complicated,
non-convex decision boundaries.
}

This case study paper
is an applied study using existing techniques (mainly Craig interpolation)
in a new practical context (XAI).
We share our lessons learned after experimenting with various real-life
benchmarks, evaluate the quality of explanations generated using different interpolation techniques and further reduced using unsatisfiable cores.
We demonstrate the benefits in carefully selected experiments with respect to the state-of-the-art
to see how the techniques apply and how different strategies can be combined.
For all kinds of benchmarks, no matter the size and the domain, the resulting explanations are more meaningful for the user.
Moreover,
computing interpolation-based explanations
scales better with the size of the input than existing methods.

\subsubsection*{Related Work}
Most of the classic approaches to XAI either perform
analysis at the \emph{unit (neuron) level}~\cite{Erhan_VisualizingDNNFeatures,Zeiler_VisualizingConvNets}, which however fails to capture global properties or interdependencies between units;
are \emph{gradient-based} methods~\cite{Simonyan_ICLR14_DeepInsideConvNets,Bach_LayerwiseRelevancePropagation}, which describe the network behavior around a particular sample;
or \emph{inversion} methods~\cite{Mahendran_CVPR2015_UnderstandingDeepImageRepresentations},
which provide global views, but are still only approximating.
There is also a range of \emph{model-agnostic} methods~\cite{Baehrens_ExplainIndividualDecisions,Lundberg_NIPS17_InterpretModelPredictions,Ribeiro_AAAI18_Anchors},
including SHAP~\cite{10.5555/3722577.3722589} and LIME~\cite{Ribeiro_SIGKDD16_WhyShouldITrustYou},
which can be applied to any classifier (not only NNs),
but often yield logically inconsistent explanations~\cite{Marques-Silva2023}.
All~\cite{%
Erhan_VisualizingDNNFeatures,Zeiler_VisualizingConvNets,%
Simonyan_ICLR14_DeepInsideConvNets,Bach_LayerwiseRelevancePropagation,%
Mahendran_CVPR2015_UnderstandingDeepImageRepresentations,%
Baehrens_ExplainIndividualDecisions,Lundberg_NIPS17_InterpretModelPredictions,Ribeiro_AAAI18_Anchors,%
10.5555/3722577.3722589,Ribeiro_SIGKDD16_WhyShouldITrustYou}
rely on approximations
or are \emph{sample-based},
that is,
the behavior outside the space of the underlying dataset
or samples generated from the model remains unknown.

We focus on formal approaches to XAI that offer \emph{strict guarantees},
exploiting computational engines like \Marabou{}~\cite{Katz_CAV19_MarabouFramework}
or SMT solvers~\cite{cvc5,z3,opensmt2,mathsat5,yices2,smtinterpol}.
State-of-the-art methods in this area address only
special cases of our concept:
constraints over individual features,
leading to interval-based explanations~\cite{Ignatiev:24}
that cannot reflect feature relationships
and hence approximate complex decision boundaries
such as in Figure~\ref{fig:bnd:real}.
Section~\ref{sec:exp} demonstrates
generalizations of the existing explanations,
resulting in strictly larger spaces.
Early methods in this domain aimed
at so-called abductive explanations~\cite{Ignatiev_2019,Shih_IJCAI18_SymbolicApproach,Seshia_TowardsVerifiedAI,Marques-Silva2023,Wu_2023,LaMalfa_IJCAI21_GuaranteedOptimalRobustExplanations}
that are primarily sample-based,
resulting in zero-volume spaces.
The term \emph{abduction}~\cite{DilligDLM13,Albarghouthi2016,Prabhu:Abduction1,Prabhu:Abduction2}
is misused,
since unlike abductive explanations,
it is not restricted to samples.
All the approaches~\cite{Ignatiev:24,Ignatiev_2019,Shih_IJCAI18_SymbolicApproach,Seshia_TowardsVerifiedAI,Marques-Silva2023,Wu_2023,LaMalfa_IJCAI21_GuaranteedOptimalRobustExplanations}
also heavily depend on selecting
a~specific order of the features in which the constraints are relaxed.
Furthermore,
they might
build an exponential number of verification queries
to an off-the-shelf NN verification engine.
In contrast,
our method directly uses
one satisfiability query to
an interpolating SMT solver
to generate an interpolant.

\subsubsection*{Acknowledgement} A similar version of this paper has been accepted to CAV 2025~\cite{spaceex}.

\section{Background}
\label{sec:background}


A classification problem~\cite{Goodfellow_DeepLearning} is concerned with mapping input data into a predefined set of classes $\mathcal{K} = \{c_1, \dots, c_K\}$.
Given a set of \emph{features} $\mathcal{F} = \{1, \dots, m\}$, each $i \in \mathcal{F} $ takes values from a~discrete or continuous \emph{domain} $\mathcal{D}_i$. 
The \emph{feature space} is defined as $\mathbb{F} = \mathcal{D}_1 \times \mathcal{D}_2 \times \dots \times \mathcal{D}_m$.
By $\mathbf{x}$ =  $(x_1, \dots, x_m) $ we denote a point in $\mathbb{F}$, where each $x_i$ is a~\emph{feature variable} (if clear from context,
called features) taking values from $\mathcal{D}_i$.
A \emph{sample point} $\Sample{} = (s_1,\dots,s_m) \in \mathbb{F}$ contains constants representing concretes value from $\mathcal{D}_i$, \dots, $\mathcal{D}_m$.
A \emph{classifier} $\mathcal{M} = (\mathcal{F}, \mathcal{D}, \mathbb{F}, \mathcal{K}, \kappa)$ contains a \textit{classification function} $\kappa : \mathbb{F} \rightarrow \mathcal{K} $. 
An \emph{instance} is a~pair $(\Sample{}, c)$, where $\Sample{} \in \mathbb{F}$  and $c \in \mathcal{K}$ is a prediction $c = \kappa(\Sample{})$.
%
    \label{def:spacecls}
    Given a~classifier
    $\mathcal{M} = (\mathcal{F}, \mathcal{D}, \mathbb{F}, \mathcal{K}, \kappa)$,
    the \emph{class space} $\ImpSpace{c} \subseteq \mathbb{F}$
    of $c \in \mathcal{K}$ is $\{ \mathbf{x} \in \mathbb{F} \mid \kappa(\mathbf{x}) = c \}$.
    A~\emph{classification rule} for $c$
    is a formula $\varphi_c$ such that $
        \forall\, \mathbf{x} \in \mathbb{F} \mathbin{.}
        \varphi_c(\mathbf{x}) \iff \kappa(\mathbf{x}) = c$.
%
    If $\kappa$ is defined on the whole feature space~$\mathbb{F}$
    then it can be partitioned as $
        \mathbb{F} = \bigcup_{c \in \mathcal{K}}\, \ImpSpace{c}$, and
    for all $a, b \in \mathcal{K}$, if $a \neq b$ then $\ImpSpace{a} \cap\, \ImpSpace{b} = \emptyset$
    (because $\kappa$ must not be ambiguous).
%
    \label{def:bnd}
    A \emph{class boundary} of $c$
    is the topological boundary of $\ImpSpace{c}$.
    The \emph{decision boundary}
    is the union of class boundaries of all classes $c \in \mathcal{K}$ in a~classifier $(\mathcal{F}, \mathcal{D}, \mathbb{F}, \mathcal{K}, \kappa)$.

\Hide{
Decision boundaries have zero volumes
if the assumptions of Corollary~\ref{cor:split_space} hold.
}

\section{Space Explanations}
\label{sec:spaceex}

Explanations produced by existing tools do not capture relations
between the input features,
and consequently, they cannot approximate the decision boundary. To address it, this paper presents a~\emph{logic}-based approach
to compute explanations with the flexibility
of representing different shapes of the feature space.
\Hide{%
For example in Figure~\ref{fig:bnd:art}, to analyze the~$\oplus$-sample points within the area $[2,4] \times [1,3]$,
it is sufficient to estimate the decision boundary and capture the behavior of the classifier there.
An explanation for the sample point at position $(2,2)$ is a polygon that estimates the shape of the decision boundary.
As a result, the polygon indeed \emph{explains} why the sample point was mapped to class~$c_1$.

because its coordinates, that is, the features,
satisfy the \emph{relationships} given by the approximation of the decision boundary,
corresponding to the shape of the polygon.
An explanation applies to multiple sample points, and the union of all three explanations
approximates the decision boundary in the area $[2,4] \times [1,3]$.
}%

\begin{definition}[Space Explanation, Impact Space]
    \label{def:spaceex}
    Given a classifier~$\mathcal{M}$
    computing a~classification function~$\kappa$
    from feature space~$\mathbb{F}$,
    and a class~$c$,
    a~\emph{space explanation} of $c$
    is a logic formula~$\varphi$ such that $
        \forall\, \mathbf{x} \in \mathbb{F} \mathbin{.}
        \varphi(\mathbf{x}) \implies \kappa(\mathbf{x}) = c$.
    The \emph{impact space} of~$\varphi$,
    $\mathbb{F}_{\varphi} \subseteq \mathbb{F}_c$,
    is a set $\{ \mathbf{x} \in \mathbb{F} \mid \varphi(\mathbf{x}) \}$.
\end{definition}

Space explanations represent
sufficient conditions of classification to class~$c$.
Since $\varphi \implies \varphi_c$,
classification rules are also space explanations.
%
%
Hence,
we have a~general concept of explanations
with the following benefits:
\begin{enumerate}
    \item The \emph{shape} of the explanations is \emph{not restricted},
    hence it is possible to approximate arbitrarily complex
    class spaces and decision boundaries.%
    \label{it:spaceex:shape}
    \item It is possible to \emph{capture relationships} among features
    and truly \emph{explain} the \emph{reasons} behind the classification.%
    \label{it:spaceex:relship}
    \item Space explanations can be merged or intersected with each other.
    \label{it:spaceex:compose}
\end{enumerate}
In 
Figure~\ref{fig:bnd:real}, for example, items~1 and~2 addressed
by accurate approximation of the decision boundary
using an explanation in the shape of a~convex polygon.

\Hide{
We need to be able to compute such rich explanations
that capture relationships between the features
even in high-dimensional spaces.
Fortunately,
Craig interpolation saves the day.
}

\Hide{
Projecting the space into lower dimensions is also not necessarily useful:
Assume that Figure~\ref{fig:bnd} is a~cut from a~three dimensional feature space.
If the space somewhere behind this cut eventually formed a~full square,
the relationship of features~$x$ and~$y$,
corresponding to the depicted decision boundary,
would remain invisible in the projection into~$x$ and~$y$,
because we would only see the full square.
}


 We assume that for all classes,
    at least one sample point is classified into the class,
    meaning that none of the classes are redundant.
Given a~formula $\psi := \psi_\mathcal{M} \land \psi_\mathcal{D} \land \neg \psi_c$ where: $\psi_\mathcal{M}$ encodes
    the neural network~$\mathcal{M}$, $\psi_\mathcal{D}$ encodes
    the domains~$\mathcal{D}$ of the feature space,
    i.e. $\mathbf{x} \in \mathbb{F}$, $\neg \psi_c$ encodes
    the constraint that the outcome of the classification
    is \emph{not} class~$c$,
it is \emph{satisfiable}
with no additional restrictions.
Now suppose a~sample point~$\Sample{}$
classified as class~$c$.
Using the encoding
\(
    \SampleExpl := \bigwedge_{i \in \mathcal{F}}\ x_i = s_i
\),
formula $\SampleExpl \land \psi$
is in turn \emph{unsatisfiable}.

%
This principle can be generalized
by exploiting the fact that
a~space explanation~$\varphi$ of class~$c$ \emph{guarantees}
classification to the class
for all points covered by the explanation,
so formula $\varphi \land \psi$ is also unsatisfiable.
Such a~formula enables the use of Craig interpolation\footnote{
Given an unsatisfiable formula $A \land B$, a \emph{Craig interpolant}~\cite{Cra57} is a formula $I$ such that $A$ implies $I$, $I \land B$ is unsatisfiable, and $I$ uses only the common variables of $A$ and $B$.
We denote interpolant~$I$ computed by an~interpolation procedure $\Itp$
from formulas~$A$ and~$B$ by $I = \Itp{}(A, B)$.
}: interpolant $I = \Itp{}(\varphi, \psi)$
is a~space explanation of~$c$
and it satisfies $\varphi \implies I$
(and hence $\ImpSpace{\varphi} \subseteq \ImpSpace{I}$).

\Hide{
\begin{proof}
    The implication $\varphi \Rightarrow I$ holds
    by the definition of Craig interpolation.
    The interpolation also guarantees that $I \land \psi$ is unsatisfiable,
    meaning that the classification cannot change,
    assuring that $I$ satisfies Definition~\ref{def:spaceex}.
    \QED
\end{proof}
}

Hence,
we have a~universal means of generalization of existing space explanations,
still guaranteeing the classification.
Although space explanations are not tied to specific data samples,
formulas~$\SampleExpl$
can be used as a~starting space explanation,
for which
the interpolants are computed quickly
because all input variables are fixed to concrete values.
This offers the following improvements over existing methods:
\begin{enumerate}
    \item The captured feature relationships
    stem from a~mathematical \emph{proof} of the classification,
    hence providing \emph{meaningful} information.%
    \label{it:itp:proof}
    \item The concept is \emph{flexible},
    offering to use \emph{arbitrary} Craig interpolation algorithms.
    The resulting explanations exhibit various logical strengths and forms.%
    \label{it:itp:flex}
    \item It is possible to further \emph{generalize}
    any \emph{existing} space explanation.%
    \label{it:itp:general}
\end{enumerate}

\Hide{
Existing proof-based interpolation algorithms
(cf. Section~\ref{sec:background:itp})
allow control over the logical strength of the resulting explanations.
Furthermore,
they may produce syntactically more general formulas
than the original explanation.
For example,
some of the interpolation algorithms
in practice transform a~sample point $\SampleExpl$
into a~space explanation
where the atoms do not only involve individual features.
Moreover,
the resulting explanations \emph{capture feature relationships}
that stem from a~resolution proof of the classification.
Hence, even if we start from~$\SampleExpl$,
we can already deliver meaningful explanations.
We confirm these claims
empirically in Example~\ref{example:spaceex:itp}
and in Section~\ref{sec:exp}.
}

Interpolation algorithms
can benefit from special cases~\cite{Gurfinkel13}
where a~formula can be partitioned.
We address just one example
when a~space explanation can be split into two parts $\varphi_A \land \varphi_B$:
given a~Craig interpolation procedure $\Itp{}$
with $I = \Itp{}(\varphi_A, \varphi_B \land \psi)$,
then
$I \land \varphi_B$ is a~logically weaker space explanation.
We used this technique
to identify pair feature relationships
in Figure~\ref{fig:bnd:real}.
\Hide{%
Unsatisfiable cores are a~special case of Craig interpolation
even within the context of space explanations:
Since any space explanation~$\varphi$ can be expressed
in an equivalent conjunctive form
\(
    \varphi'(\mathbf{x}) =
    \bigwedge_{i \in \mathcal{I}}\, \varphi_i(\mathbf{x})
\)
(e.g. with $\mathcal{I} = \{1\}$ and $\varphi_1 = \varphi$),

we can compute the unsatisfiable core $\Ucore{}(\varphi, \psi)$
which is guaranteed to have non-strictly larger impact space
and
possibly reduced formula\Hide{\footnote{%
The same principle is used for example
within abductive explanations~\cite{Ignatiev_AAAI19_AbductionBasedExplanations}.
However,
here we do not put any restrictions on the structure of the formulas.
}}.
}%

\begin{figure}[t!]
    \centering
    \includegraphics[trim={1.6cm 0cm 1.6cm 0.5cm},clip,width=0.93\textwidth]{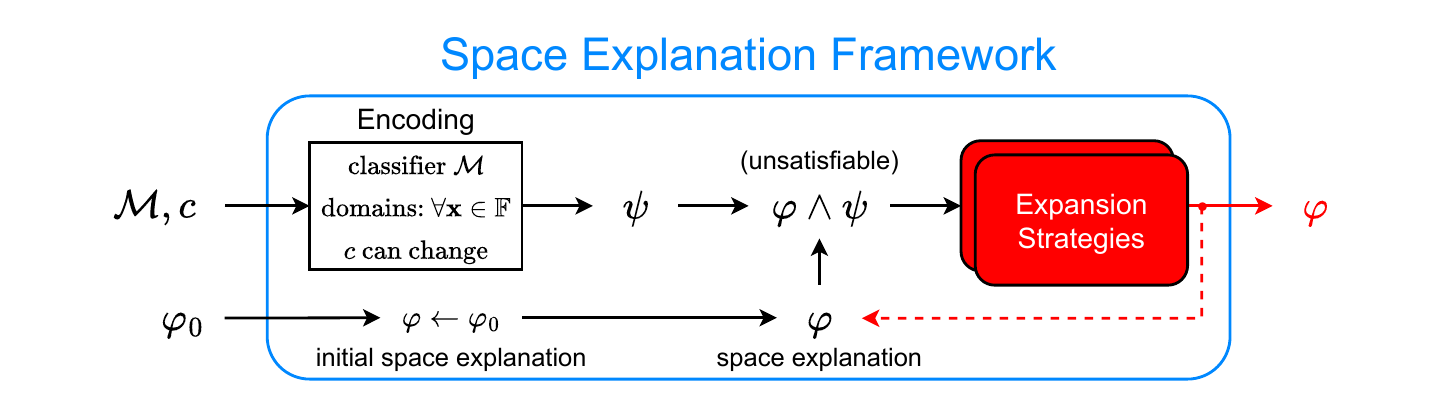}
    \caption{Algorithmic framework: starting from an initial space explanation~$\varphi_0$, the impact space is expanded and a logically weaker explanation~$\varphi$ is produced. The user may select from multiple strategies for further optimization (cf. the dashed arrow).}
    \label{fig:framework}
\end{figure}

Figure~\ref{fig:framework} gives a~high-level representation
of a~fully functioning and flexible algorithmic \emph{framework}
that provides expansion of space explanations
using a~collection of \emph{strategies} listed below.

\newcommand*{\Strategy}[1]{\textbf{#1}}
\newcommand*{\G}{\Strategy{G}}
\newcommand*{\R}{\Strategy{R}}
\newcommand*{\Rmin}{\ensuremath{\R{}_\mathit{min}}}
\newcommand*{\C}{\Strategy{C}}

\newcommand*{\I}{\Strategy{I}}
\newcommand*{\A}{\Strategy{A}}

\begin{itemize}
\item\textbf{Generalize (\G{}).}
Compute Craig interpolants
on an arbitrary space explanation.
%
We use the algorithms based on Farkas' lemma ($\ItpF$)
or their logically stronger \emph{decomposing} variants~\cite{Blicha19} ($\ItpDF$),
the dual versions of these algorithms ($\ItpF'$, $\ItpDF'$) to yield weaker interpolants~\cite{Silva10},
and an algorithm~\cite{Alt17}
($\ItpFactor$)
parametrized by a~rational factor~$f \in [0,1]$
with the logical strength in between $\ItpF$ and $\ItpF'$.
These arithmetic interpolation algorithms are combined with McMillan's propositional interpolation algorithm~\cite{McMillan03}.


\item\textbf{Reduce (\R{}).}
Weaken the formula and reduce size by computing an unsatisfiable core.
To get irreducible explanations (\Rmin{}),
we use exhaustive minimization
which may introduce significant overhead.

\item\textbf{Capture (\C{}).}
Partition an interval-like formula (e.g. a~sample) and \textbf{Generalize}.
Weaken only part~$\varphi_A$
with constraints over at least one of the given features,
hence capturing their mutual relationships.
\end{itemize}
Strategy \textbf{Generalize} and \textbf{Capture}
is often followed by \textbf{Reduce}
to simplify the formulas from unnecessary constraints.

\NewDocumentCommand{\percent}{}{%
  \texttt{\%}%
}
\NewDocumentCommand{\num}{}{%
  \texttt{\#}%
}

\newcommand*{\aItpAlg}[1]{\texttt{#1}}
\newcommand*{\astronger}{\aItpAlg{stronger}}
\newcommand*{\astrong}{\aItpAlg{strong}}
\newcommand*{\amid}{\aItpAlg{mid}}
\newcommand*{\aweak}{\aItpAlg{weak}}
\newcommand*{\aweaker}{\aItpAlg{weaker}}

\section{Experimental Evaluation}
\label{sec:exp}
We showcase\footnote{
    All table contents are reproducible:
    \url{https://doi.org/10.5281/zenodo.15490124}
}
the computation of space explanations
for selected trained neural networks
and demonstrate that they are
substantially more expressive and meaningful
since our approach
captures feature relationships and approximates decision boundaries.
We implemented the novel Space Explanation Framework
in a prototype tool
\spexplain{}\footnote{
    \url{https://github.com/usi-verification-and-security/spexplain}
}, focusing on \LRA{} logic, 
on top of the \Opensmt{} solver~\cite{opensmt,opensmt2}%
\Hide{
\footnote{
    \Opensmt{} is the award-winning solver
    in the \LRA{} logic in all tracks,
    including the Unsat Core Track,
    in SMT-COMP 2024 (\url{https://smt-comp.github.io/2024}).
}
}
which is an interpolating solver that comes with a set of techniques assembled into the integral framework, combining SMT solving with the computation of the Craig interpolants of various size and strength~\cite{DBLP:conf/memics/HyvarinenAS15, DBLP:conf/lpar/RolliniAFHS13, DBLP:conf/vstte/AltFHS15,DBLP:conf/iccad/BruttomessoRST10,Blicha19,Alt17}.
We evaluated
the following NN datasets and models:

\begin{itemize}
    \item Heart attack dataset~\cite{heart_disease_45}
    focuses on predicting the risk of heart attacks
    based on various medical indicators of patients,
    it contains 13 input features and 2 possible classification outcomes:
    high or low risk.
    We trained the NN using one hidden layer with 50 neurons,
    and used a dataset with 303 sample points.
    \item Obesity dataset~\cite{obesity}
    provides data for estimating obesity levels in individuals
    based on their eating habits and physical condition,
    resulting in 15 input features and 7 classes.
    We trained the model using 3 hidden layers
    with 10, 20, and 10 neurons, respectively,
    and used a dataset with 50 sample points.
    \item MNIST dataset~\cite{mnist}
    is a collection of grayscale images of handwritten digits (0–9)
    with 784 inputs,
    commonly used within image classification tasks
    as a~reference for
    training, evaluation, and verification of machine learning models.
    We trained the model using 784 inputs and a~hidden layer with 200 neurons,
    and used a dataset with 50 sample points.
\end{itemize}
Instances of interpolation algorithms
use the notation
$\ItpDF \mapsto \astronger$,
$\ItpF \mapsto \astrong$,
$\ItpF' \mapsto \aweak$,
$\ItpDF' \mapsto \aweaker$,
and
$\ItpFactor \mapsto \amid{}$
with
$f := 0.5$.
%
The evaluation
uses 2~hour \emph{time-out} to compute all explanations
for a~dataset and an explanation setup,
and was performed on a Linux 5.4 machine
with 256~GB physical memory
and
AMD\textsuperscript{\textregistered} EPYC 7452 32-core CPU.

\begin{figure}[!t]
    \centering
\begin{subfigure}{\textwidth}
    \includegraphics[width=\textwidth]{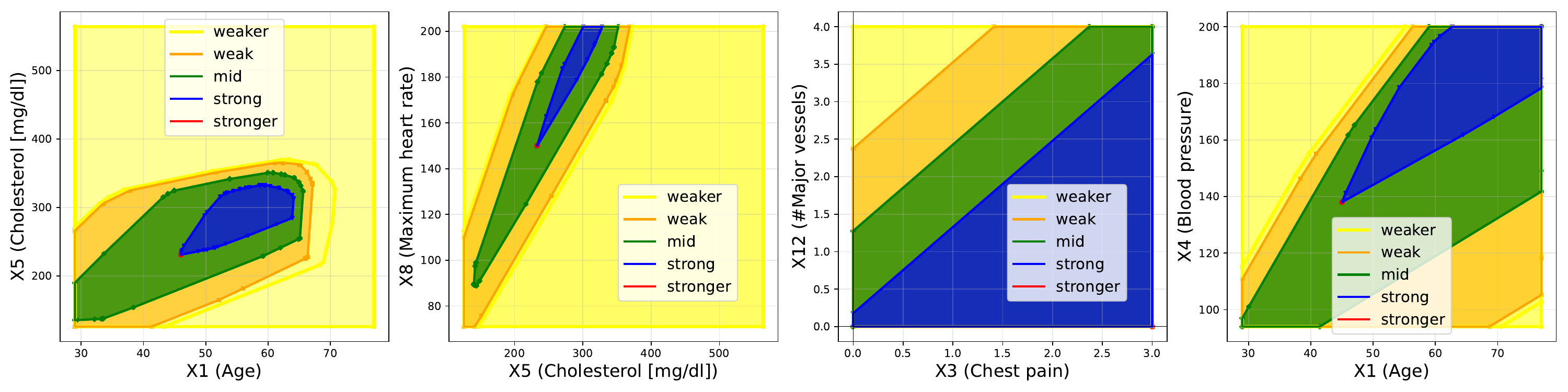}
    \caption{\textbf{Generalize} (\G{}): Interpolation-based explanation impact spaces (projections)}
    \label{fig:exp:itps}
\end{subfigure}
\begin{subfigure}{\textwidth}
    \includegraphics[width=\linewidth]{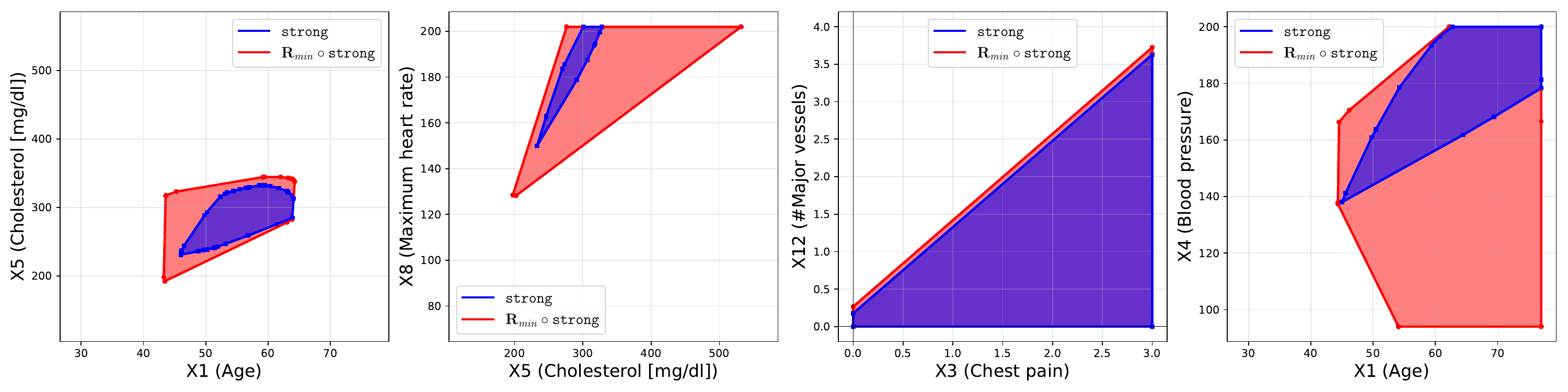}
    \caption{\textbf{Reduce} ($\Rmin{} \circ \G{}$): \G{}-\astrong{} explanations and their exhaustive simplifications (projections)}
    \label{fig:exp:ucore}
\end{subfigure}
\caption{Projections into selected pairs of features
of interpolation-based explanations (\G{})
of the heart-attack risk model,
possibly combined with reduction (\Rmin{}).
}
\label{fig:exp:itps:merged}
\end{figure}

\subsubsection*{Evaluation of Strategies}
Instantiations of strategy \textbf{Generalize} (\G{})
provide flexibility of covering spaces with various size,
illustrated in
Figure~\ref{fig:exp:itps}
on
selected sample points from the dataset.
We resolve the issue of visualizing high-dimensional spaces
by selecting candidate pairs of features
from the heart attack dataset
and plotting the projection\footnote{
    \label{footnote:projection}%
    Projections
    offer a visual intuitive understanding of the geometrical differences
    between impact spaces.
    While not capturing the complete information contained
    in higher-dimensional spaces,
    they are still more informative than simple slices.
}
of the impact spaces
into two dimensions.
The examples reveal the similarity of feature relations
despite aiming at different logical strengths.
However,
we spot that
\aweaker{} here always covers the whole projection\footnote{
    Notably, the explanation does not cover the whole feature space.
    However, high-dimensional projections (e.g., from 13D to 2D) collapse information
    from other dimensions, making explanations appear larger.
},
while
\astronger{} in turn results just in the original sample point.
This is no exception:
Table~\ref{tab:itp:stats}
reports
that \astronger{} relaxes no features at all
(i.e. each feature is still fixed to the original value),
producing way too strong interpolants.
Algorithm \astrong{} sometimes does not relax all features.
Next,
the table
reports the average size of the resulting formulas term-wise
(the number of in/equalities).
The average runtimes per sample point
are rather small
and do not vary among the approaches,
but are sensitive to more complex models.
\Hide{
To conclude,
we showed that several possible algorithms
allow to explore spaces of different size
yet often still resembling similar shapes.
}

\setlength{\tabcolsep}{0.5pt}

\begin{table}[!t]
\centering
\caption{Average performance of strategy \G{} using different $\Itp{}$ algorithms.}
\label{tab:itp:stats}
\begin{tabular}{l|ccc|ccc|ccc|}
\cline{2-10}
 & \multicolumn{3}{c|}{Heart attack} & \multicolumn{3}{c|}{Obesity} & \multicolumn{3}{c|}{MNIST} \\
\multicolumn{1}{l|}{$\Itp{}$ algorithm} & relaxed &  \verb|#|terms & time[s] & relaxed & \verb|#|terms & time[s] & relaxed & \verb|#|terms & time[s] \\ \hline
\multicolumn{1}{|l|}{\astronger} & 0\%  & 20.1  & 0.03 & 0\% & 29.0 & 0.30 & 0\%      & 927.3  &9.17 \\
\multicolumn{1}{|l|}{\astrong} & 97\% & 51.0  & 0.04 & 72\% & 45.8 & 0.30 & 100\% & 209.0 &10.12 \\
\multicolumn{1}{|l|}{\amid} & 100\%     & 51.0  & 0.04 & 100\% & 45.8 & 0.34 & 100\% & 209.0  &10.25 \\
\multicolumn{1}{|l|}{\aweak} & 100\%    & 51.0  & 0.04 & 100\% & 46.1 & 0.30 & 100\% & 209.0  &10.42 \\
\multicolumn{1}{|l|}{\aweaker} & 100\%  & 198.2 & 0.04 & 100\% & 64.1 & 0.42 & 100\% & 67330.6  &10.55 \\
\hline
\end{tabular}
\end{table}

\Hide{
For instance, a strong interpolation algorithm such as Dual Farkas interplant with Dual McMillan algorithm, which produces disjunctions of formulas, produces explanations whose projections span the entire 2D space.
In contrast,
a strong one like Decomposing Farkas interpolant with McMillan algorithm
exhibits no generalization.
}


\newcommand*{\NC}{\textbf{NC}}

Strategy \textbf{Reduce} (\R{}, \Rmin{})
introduces
a trade-off between formula simplification
and runtime overhead
when applied on top of \textbf{Generalize}
(e.g., $\R{} \circ \G{}$),
using
an unsatisfiable core (\R{})
or an irreducible one (\Rmin{}).
Figure~\ref{fig:exp:itps}
demonstrates side by side with Figure~\ref{fig:exp:ucore}
that the projection of \astrong{}
is further expanded by~\Rmin{}
(\R{} expands none of the four presented \astrong{} projections),
but in different directions compared to weaker algorithms.
We confirmed this phenomenon by running extensive subset-comparisons
which in cases such as $(\astrong{}, \Rmin{})$ vs. $(\aweak{}, -)$
often yielded not comparable (\NC) results:
the spaces intersect but none of them subsumes the other.
Table~\ref{tab:ucore} shows
that \Rmin{} significantly reduces the formula
but with a~high cost,
while \R{} offers a~convenient balance. 
The reduction times-out (X) with more complex models.

\begin{table}[!tb]
\centering
\caption{Time overhead and impact of \textbf{Reduce} on top of \textbf{Generalize} ($\R{} \circ \G{}$).}
\label{tab:ucore}
\begin{tabular}{c|l|ccc|ccc|}
\cline{2-8}
 & \multicolumn{1}{c|}{\G{}} & \multicolumn{3}{c|}{\num terms} & \multicolumn{3}{c|}{time [s]} \\
 \multirow{7}{*}{\rotatebox{90}{ \scriptsize Heart attack}} & $\Itp$ algorithm & $-$ & \R{} & \Rmin{} & $-$ & \R{} & \Rmin{} \\
\hline
&  \astronger{}   & 20.1 &  20.1 & 9.5 & 0.03 & 0.03 & 3.14 \\
&  \astrong{}     & 51.0 &  44.6 & 3.8 & 0.04 & 0.83 & 16.77 \\
&  \amid{}        & 51.0 &  39.4 & 4.6 & 0.04 & 0.66 & 11.80 \\
&  \aweak{}       & 51.0 &  38.8 & 6.4 & 0.04 & 0.56 & 7.95 \\
&  \aweaker{}     & 198.2 & 197.9 & 25.9 & 0.04 & 0.08 & 7.14 \\
\hline
\multirow{5}{*}{\rotatebox{90}{ \scriptsize Obesity}}
& \astronger{}    & 29.0 & 29.0  & X & 0.30 & 0.38 & X  \\
& \astrong{}      & 45.8 & 41.7  & X & 0.30 & 7.94 & X \\
& \amid{}         & 45.8 & 42.7  & X & 0.34 & 20.62 & X \\
& \aweak{}        & 46.1 & 41.7  & X & 0.30 & 35.37 & X \\
&  \aweaker{}     & 64.1 & 58.0  & X & 0.42 & 23.44 &  X \\
\hline
\multirow{5}{*}{\rotatebox{90}{ \scriptsize MNIST}}
& \astronger{}  & 927.3 & 927.3  & X &  9.17 & 10.45 & X \\
& \astrong{}   &  209.0 & X  &  X    & 10.12 & X & X \\
& \amid{}      &  209.0 & X  &  X    & 10.25 & X & X \\
& \aweak{}     &  209.0 & X  &  X    & 10.42 & X & X \\
&  \aweaker{}  &  67330.6 & X & X & 10.55 & X & X \\
\cline{2-8}
\end{tabular}
\end{table}

\begin{figure}[!tb]
    \centering
    \begin{subfigure}{\textwidth}
        \centering
        \includegraphics[width=\textwidth]{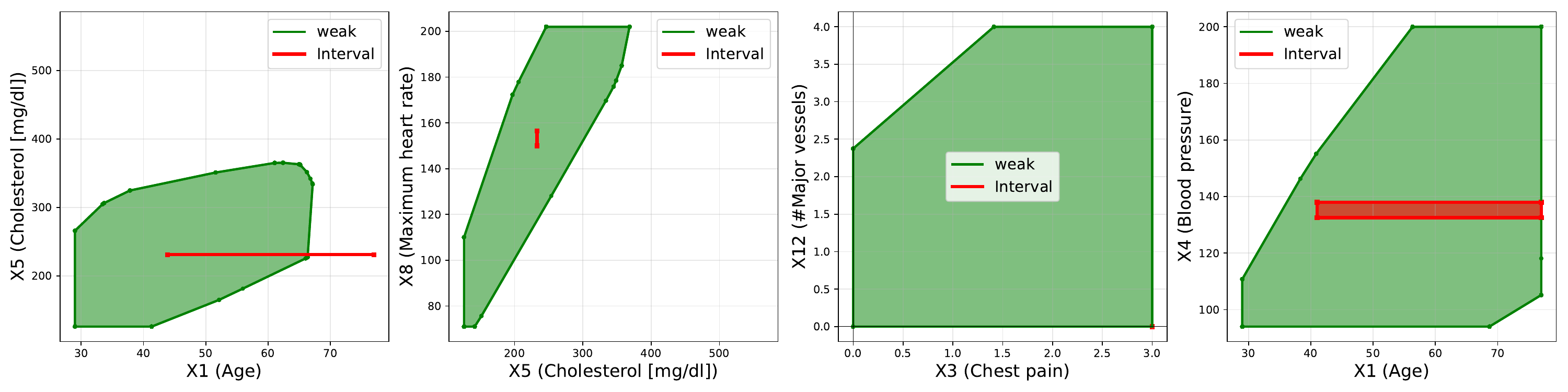}
        \caption{Projections of \G{} explanations (green) compared to $\I{} \circ \A{}$ explanations (red)}
        \label{fig:exp:itpvsinterval-based}
    \end{subfigure}
    \vspace{0.2cm}
    \begin{subfigure}{\textwidth}
        \centering
        \includegraphics[width=\textwidth]{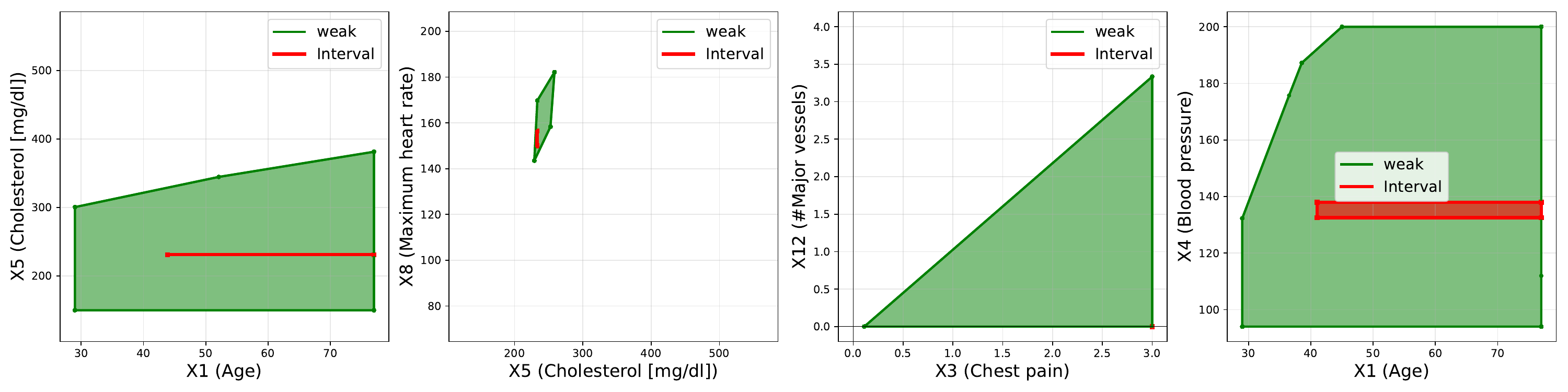}
        \caption{Projections of \G{} explanations (green) on top of $\I{} \circ \A{}$ explanations (red), resulting in $\G{} \circ \I{} \circ \A{}$}
        \label{fig:exp:generalize}
    \end{subfigure}

    \caption{Comparing projections of interpolation-based (\textbf{Generalize}, \G{})
    and interval explanations ($\I{} \circ \A{}$)
    into selected pairs of features
    of the heart-attack risk model.}
    \label{fig:exp:generalize:merged}
\end{figure}

Figure~\ref{fig:exp:itpvsinterval-based}
follows with a~comparison of projections
between strategy~\G{}
(cf. Figure~\ref{fig:exp:itps}),
specifically selecting \aweak{} algorithm
that produces large yet convex impact spaces,
and an approach
that is based on intervals,
denoted as~\I{},
that computes the explanations
on top of an irreducible abductive explanation,
denoted by~\A{},
resulting in $\I{} \circ \A{}$.
Since the state-of-the-art-approach~\cite{Ignatiev:24}
is implemented for decision trees and not for NNs,
we implemented the specialized computation of interval explanations
using the \Marabou{} verifier
and a~strict limit on the number of attempts of particular relaxations.
\Hide{%
However,
we did not succeed at implementing the classification into multiple classes
that is necessary for the Obesity and MNIST datasets.
}%
We concentrate on comparison using the heart attack model.
The projections of interval-based explanations form rectangles,
lines, or even single points due to the lack of their expressivity.
In contrast,
\G{} explanations cover larger areas
with less limited shapes,
yet do not entirely subsume interval explanations:
Even if a projection is subsumed,
it does not mean that the space is subsumed in the other dimensions as well.
While the intervals are being relaxed in a~certain order,
hence strictly preferring certain directions over others,
the interpolation aims at more general expansions. 
Furthermore,
the expansion
of the interval explanations is limited
by decision boundaries
in the other dimensions.
This observation is confirmed
by pairwise subset-comparisons,
consistently arriving at \NC{} results.

\begin{table}[!tb]
\centering
\setlength{\tabcolsep}{3.1pt}
\caption{Average performance of \G{} vs. \A{} and \I{}
strategies
in heart-attack model.}
\label{tab:stats_itpVSint}

\begin{tabular}{l!{\vrule width 1pt}c c c c!{\vrule width 1pt}}
 \cline{2-5}
 & relaxed & \num terms &\num solver calls & time [s] \\ \hline
\multicolumn{1}{|l!{\vrule width 1pt}}{\A{}} & 38\% & 8.1  & 13 & 0.08 \\
\multicolumn{1}{|l!{\vrule width 1pt}}{$\I{} \circ \A{}$} & 79\% & 9.3  & 40.4 & 0.53 \\
\multicolumn{1}{|l!{\vrule width 1pt}}{\G{}} & 100\% & 51.0 & 1 & 0.04  \\
\multicolumn{1}{|l!{\vrule width 1pt}}{$\G{} \circ \A{}$} & 100\% & 45.3 & 1 & 0.39 \\
\multicolumn{1}{|l!{\vrule width 1pt}}{$\G{} \circ \I{} \circ \A{}$} & 100\% & 63.5 & 1 & 2.53 \\
\hline
\end{tabular}
\end{table}

Table~\ref{tab:stats_itpVSint} gives
comparison of the average performance of the approaches
(similarly as in Table~\ref{tab:itp:stats}),
including irreducible abductive explanations (\A{}).
\G{} explanations (row~\#3) are computed
even faster than \A{} explanations,
requiring just one call to the logical solver.
Interval explanations did not always succeed in relaxing all features.
Nevertheless,
the formulas are simpler when based on intervals or samples.
Formulas from~\G{} could be reduced but would exceed the computation time of intervals.
The observations
remain similar when using other algorithms than \aweak{}.
Figure~\ref{fig:exp:generalize}
further illustrates
the potentials of strategy \G{}
when running
on top of arbitrary existing explanations,
such as interval explanations
(i.e., $\G{} \circ \I{} \circ \A{}$),
yielding strictly larger spaces
(confirmed by subset-comparison checks).
Nonetheless,
the projections of the interpolants
in Figures~\ref{fig:exp:itpvsinterval-based} vs.~\ref{fig:exp:generalize}
(i.e., \G{} vs. $\G{} \circ \I{} \circ \A{}$)
sometimes differ
due to more guided constraints induced by the intervals.
Next,
we revisit Table~\ref{tab:stats_itpVSint}
on rows~\#4--5,
including $\G{} \circ \A{}$ as well.
The observations remain similar
except that
when generalizing on top of a~more general starting point,
the runtime increases.
Still, our generalization of \A{} explanations
runs faster than if
using intervals
(i.e., $\I{} \circ \A{}$ vs. $\G{} \circ \A{}$).

\begin{figure}[!tb]
    \centering
    \includegraphics[width=\linewidth]{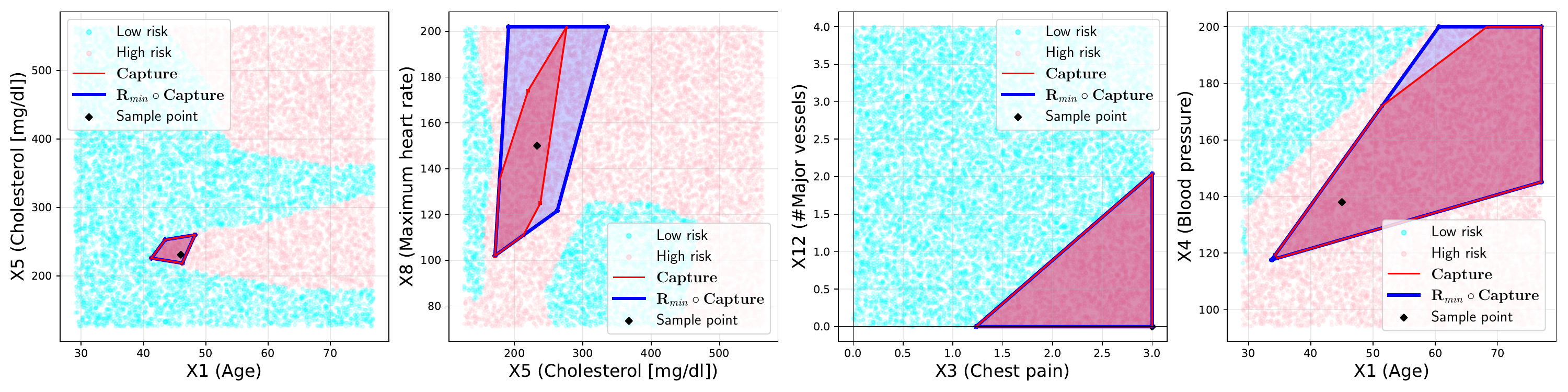}
    \caption{Approximation of decision boundaries and feature relations using strategies~\C{} and $\Rmin{} \circ \C{}$ within selected pairs of features of the heart-attack risk model.}

    \label{fig:exp:db}
\end{figure}

Strategy \textbf{Capture} (\C{})
directly aims at identifying relationships among selected features
and consequently at approximating decision boundaries.
The generalization is guided
because
it focuses on selected dimensions
and picks a~slice of the feature space,
leaving all other features fixed to the original values.
Although such limited exploration
does not reveal anything about other dimensions, 
it extracts partial information
helpful to understand the classification.
Moreover,
it is easy to sample two or three-dimensional slices
and compare the explanations with estimated class spaces
and decision boundaries.
Figure~\ref{fig:exp:db}
shows the explanations
of the sample points and feature pairs
as in Figures~\ref{fig:exp:itps:merged} and~\ref{fig:exp:generalize:merged},
this time using \C{} instead of \G{},
sticking to algorithm \aweak{}.
The interpolation captures even non-trivial decision boundaries
and
some of the relations intuitively resemble real-world phenomena.
For example,
the plot on the right identifies a~high risk of heart attack
according to increasing age and decreasing blood pressure.

\begin{table}[!tb]
\centering
\caption{Quantitative subset-comparison
of
\C{}
vs.
\G{} explanations,
combined with \R{} or \Rmin{},
ranging over selected features
where the others are fixed to the original values\Hide{ of the sample}.}
\label{tab:itp:capture:pair}
\begin{tabular}{|c|rcr|ccc c|}
\hline
Features & \multicolumn{1}{c}{\C{}} & & \multicolumn{1}{c|}{$\G{}$ (sliced)} & $\supset$ & $=$ & $\subset$ & \NC{} \\
\hline
\multirow{2}{*}{$x_1, x_5$}
& $\R{} \circ \C{}$       & vs. & $\R{} \circ \G{}$  &  8\% & 43\% & 41\% &  7\% \\
& $\Rmin{} \circ \C{}$ & vs. & $\Rmin{} \circ \G{}$ & 65\% &  0\% &  0\% & 35\% \\
\Hide{
\hline
\multirow{2}{*}{$x_5, x_8$}
& $\R{} \circ \C{}$       & vs. & $\R{} \circ \G{}$  &  8\% & 43\% & 43\% &  6\% \\
& $\Rmin{} \circ \C{}$ & vs. & $\Rmin{} \circ \G{}$ & 61\% &  0\% &  0\% & 39\% \\
}
\hline
\multirow{2}{*}{$x_3, x_{12}$}
& $\R{} \circ \C{}$       & vs. & $\R{} \circ \G{}$  &  2\% & 92\% &  5\% &  1\% \\
& $\Rmin{} \circ \C{}$ & vs. & $\Rmin{} \circ \G{}$ & 89\% &  5\% &  0\% &  6\% \\
\hline
\multirow{2}{*}{$x_1, x_4$}
& $\R{} \circ \C{}$       & vs. & $\R{} \circ \G{}$  &  9\% & 54\% & 31\% &  6\% \\
& $\Rmin{} \circ \C{}$ & vs. & $\Rmin{} \circ \G{}$ & 84\% &  0\% &  0\% & 16\% \\
\hline
\end{tabular}
\end{table}

\begin{table}[!tb]
\centering
\scriptsize
\caption{Average performance of strategies \C{} and \G{}
from Table~\ref{tab:itp:capture:pair}
possibly in combination with \textbf{Reduce} (\R{} or \Rmin{}).}
\label{tab:capture}
\begin{tabular}{l|cc|cc !{\vrule width 1.5pt} cc|cc !{\vrule width 1.5pt} cc|cc|}
\cline{2-13}
& \multicolumn{4}{c!{\vrule width 1.5pt}}{Heart attack} & \multicolumn{4}{c!{\vrule width 1.5pt}}{Obesity} & \multicolumn{4}{c|}{MNIST} \\
\cline{2-13}
& \multicolumn{2}{c|}{\C{}} & \multicolumn{2}{c!{\vrule width 1.5pt}}{\G{} (sliced)} & \multicolumn{2}{c|}{\C{}} & \multicolumn{2}{c!{\vrule width 1.5pt}}{\G{} (sliced)} & \multicolumn{2}{c|}{\C{}} & \multicolumn{2}{c|}{\G{} (sliced)} \\
\multicolumn{1}{l|}{\textbf{Reduce}} & \num terms & time[s] & \num terms & time[s] & \num terms & time[s] & \num terms & time[s] & \num terms & time[s] & \num terms & time[s]\!\\
\hline
\multicolumn{1}{|l|}{$-$}     & 61.2 & 0.04 & 62.0 & 0.04 & 59.1 & 0.33 & 59.1 & 0.31 & 938.9 & 9.65 & 991.0 & 10.54 \\
\multicolumn{1}{|l|}{\R{}}    & 47.0 & 0.09 & 49.8 & 0.56 & 41.9 & 1.15 & 54.7 & 35.3 & 834.7 & 47.1 &     X &     X \\
\multicolumn{1}{|l|}{\Rmin{}} &  9.4 & 2.52 & 17.4 & 7.94 &    X &    X &    X &    X &     X &    X &     X &     X \\  
\hline
\end{tabular}
\end{table}

Using \textbf{Reduce} (\R{}, \Rmin{})
is especially useful for explanations
produced by strategy~\C{}
to enhance their interpretability.
Furthermore,
the reduction is often more efficient
than when applied to~\G{}.
Table~\ref{tab:itp:capture:pair}
shows
the quantitative subset-comparison,
that is,
how many explanations in percentage exhibited the relation
superset ($\supset$),
equivalent ($=$),
subset ($\subset$),
or not comparable (\NC{}),
between
\C{} and \G{} explanations
that have been reduced
and
while
focusing just on the slice of selected pairs of features
(cf. Figures~\ref{fig:exp:itps:merged}--\ref{fig:exp:db}).
With no reduction,
the slices are equivalent.
Yet, the formulas differ,
because
strategy~\C{}
separates the fixed features from the focused ones\footnote{
    Example:
    \G{} sliced to $x_1, x_2$
    yields
    \(
    \big(
        {2x_1 + x_3 \geq 7}
        \land
        {x_1 - x_2 + x_3 \geq 2}
    \big)
    \land
    {x_3 = 1}
    \)
    which is equivalent to
    \(
    \big(
        {x_1 \geq 3}
        \land
        {x_1 - x_2 \geq 1}
    \big)
    \land
    {x_3 = 1}
    \)
    produced by
    \C{} of $x_1, x_2$.
}.
Consequently,
the opportunities for reductions
offered by the structure of the formulas
are different.
While
strategy~\R{} usually expands the explanations better
when applied to~\G{} (i.e. $\R{} \circ \G{}$),
$\Rmin{} \circ \C{}$ always produced a~larger space within the slice
than $\Rmin{} \circ \G{}$,
if excluding \NC{} cases.
The results for $x_5, x_8$ are almost the same as for $x_1, x_5$.
Finally,
Table~\ref{tab:capture} shows
that the simplification
is more efficient
in terms of runtime and \num{}terms
when using the focused strategy \C{}
than if interpolating the whole formula with \G{}.


\subsubsection*{Scalability}
\begin{figure}[!tb]
    \begin{subfigure}{0.49\textwidth}
        \centering
        \includegraphics[trim={0cm 0.0cm 0cm 0.0cm},clip,width=\textwidth]{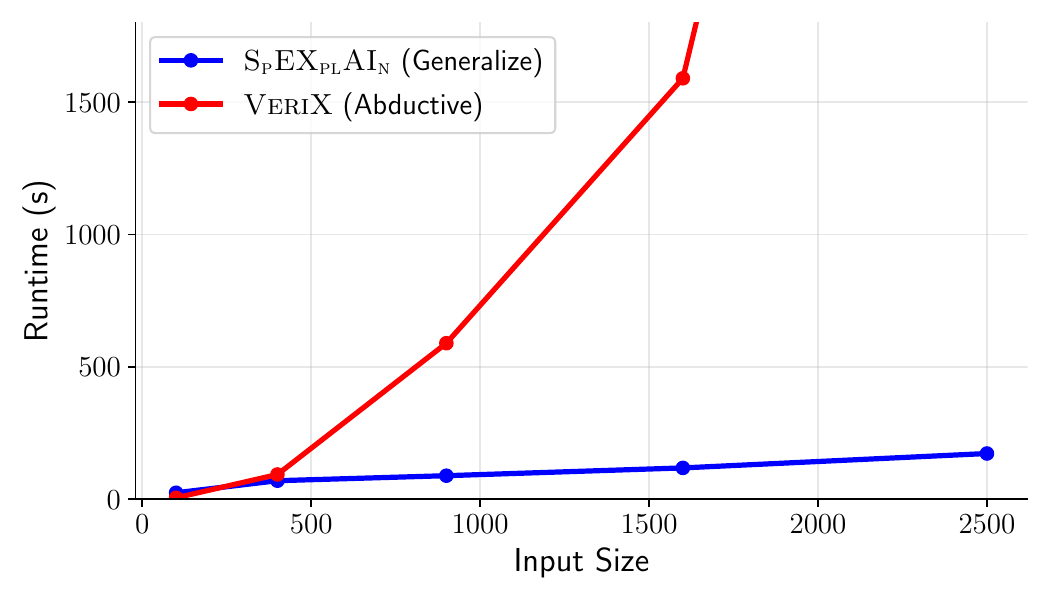}
        \caption{Scalability with input size}
        \label{fig:scale:input}
    \end{subfigure}
    \begin{subfigure}{0.49\textwidth}
        \centering
        \includegraphics[trim={0cm 0.0cm 0cm 0.0cm},clip,width=\textwidth]{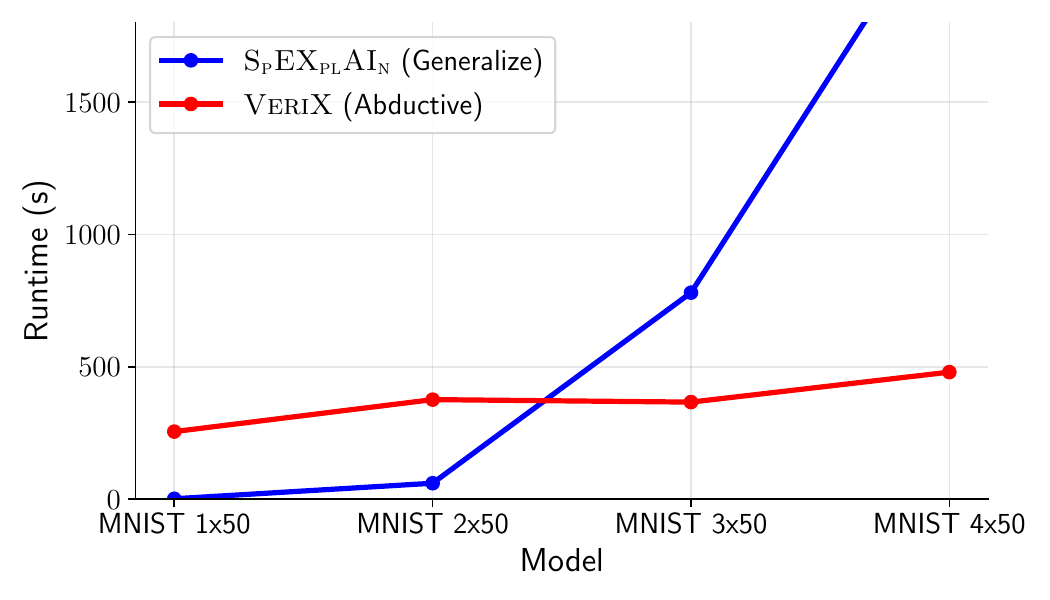}
        \caption{Scalability with NN depth}
        \label{fig:scale:layers}
    \end{subfigure}
    \caption{Scalability comparison of \spexplain{} and \verix{} on MNIST}
    \label{fig:scale}
\end{figure}

We conducted two experiments to evaluate how our prototype tool \spexplain{} scales with increasing input size
and depth of the NN
using MNIST.
We compare our method with \verix{}~\cite{Wu_2023}, a state-of-the-art abductive explanation algorithm\footnote{\url{https://github.com/NeuralNetworkVerification/VeriX}, commit b6b2cc0}.
While the approaches generate different types of explanations and are not directly comparable,
their relative scalability can still be compared.

The first experiment shows how the runtime grows with input size.
For that, we used a fixed NN architecture with two hidden layers of size 50\footnote{A smaller network was used to reduce the number of time-outs.},
and resized MNIST images to input dimensions $10 \times 10$, $20 \times 20$, $30 \times 30$, $40 \times 40$, and $50 \times 50$.
Figure \ref{fig:scale:input} shows how the runtime grows with the input size
for computing \G{} explanations and abductive explanations:
the runtime of \verix{} increases steeply,
while \textbf{Generalize} remains relatively stable.
The difference is caused by the fact that
\verix{} performs one verification query per feature, while our method requires only a single query to generate an explanation.

The next experiment examines scalability with respect to model depth.
We trained NNs with~1 to~4 hidden layers, each having 50 neurons.
Figure~\ref{fig:scale:layers} shows that
the runtime of \verix{} grows moderately with the number of layers,
while the runtime of \textbf{Generalize} increases more significantly
with deeper networks.
This is likely because, unlike \verix, our prototype tool does not apply any NN-specific optimizations within its underlying constraint solver.
Addressing this limitation is left for future work.

\Hide{
To summarize, existing approaches like \verix{} are highly sensitive to input size
while our framework is more sensitive to model depth.
Addressing this limitation is left for future work.
}

\Hide{
we conducted an experiment using a fixed neural network architecture on resized MNIST images with input dimensions of 10×10, 20×20, 30×30, 40×40, and 50×50 to show how the it scales when the input size grows.
We compared the scalability of our framework with the scalability of a state-of-the-art abductive explanation algorithm, \verix{} \cite{Wu_2023}, which is implemented in python \footnote{\url{https://github.com/NeuralNetworkVerification/VeriX}, commit b6b2cc0
}.
Also, computing inflated explanations \cite{Ignatiev:24} is strictly slower than \verix{} because they are built on top of an abductive explanation.
Note that these two methods compute different explanations, and they are not naturally comparable; we only compare how they scale.

As shown in Figure~\ref{fig:scalability}, the runtime of \verix{} increases steeply with input size, while Generalize remains relatively stable, indicating superior scalability.
 This is because unlike \verix, which issues one query per feature, our framework requires only a single query for computing the explanation.
Also, computing inflated explanations \cite{Ignatiev:24} is strictly slower than \verix{} because they are built on top of an abductive explanation.
}

\def\LLObservationWidth{0.7\linewidth}
\def\LLSolutionWidth{0.3\linewidth}

\newcommand{\LLProblem}[2][Problem]{%
\begin{tabular}{| p{\LLObservationWidth} | p{\LLSolutionWidth} |}
\hline

\multicolumn{2}{|l|}{%
\textbf{#1}:
#2%
}\\\hline
\vspace*{-1em}
\begin{enumerate}
}

\newcommand{\LLObservation}[2][]{%
\item
#2%
}

\newcommand{\LLSolution}[1]{%
\end{enumerate}
&
\vfill
#1%
\\\hline
\end{tabular}
}

\newcommand{\LLSeparator}{}

\subsubsection*{Lessons Learned}
\label{sec:lessons}

The following table summarizes our observations
regarding
the
computation,
interpretation,
simplification,
and
comparison
of explanations.
\begin{center}
\scriptsize
\begin{tabular}{| m{\LLObservationWidth} | m{\LLSolutionWidth} |}
\hline
\textbf{Observations} & \textbf{Solution}
\\\hline
\end{tabular}

\nopagebreak

\Hide{
\LLProblem{
Verification-based approaches to XAI tend to suffer from exponential number of verification queries.
}
\LLObservation[1]{Craig interpolation is fast and uses just one satisfiability query of the SMT solver.}
\LLObservation[2]{SMT solvers are highly engineered tools, although not optimized for neural networks.}
\LLSolution{Use an interpolating SMT solver for XAI. In \Hide{the }future, optimize for NNs.}

\LLSeparator
}

\Hide{
\LLProblem{How to use a space explanation?}
\LLObservation[1]{Although the process is currently rather interactive than automated, it still yields valid rules.}
\LLObservation[2]{Some explanations \Hide{are intuitive and }resemble real-world phenomena.}
\LLSolution{Domain experts (e.g., medics) can do sanity checks
and use explanations inconsistent with their knowledge
as a lead to retrain the model.}

\LLSeparator
}

\LLProblem{Explanations produced by \textbf{Generalize} are too complicated to read.}
\LLObservation[1]{\texttt{simplify} command of \textsc{Z3} or \textsc{cvc5} SMT solvers is insufficient.}
\LLObservation[2]{%
\textbf{Reduce} simplifies conjunctive formulas,
often surprisingly well.}
\LLObservation[3]{McMillan interpolation algorithm extends formulas into conjunctions and its dual into disjunctions.}
\LLSolution{Use the non-dual algorithm and simplify the explanations using \textbf{Reduce} if needed.}

\LLSeparator

\LLProblem{Information regarding just a few features is difficult to extract, even if we use \textbf{Reduce}.\!}
\LLObservation{Partitioning interval-like formulas separates selected features.}
\LLSolution{Use \textbf{Capture} and \textbf{Reduce}.}

\LLSeparator

\LLProblem{It is not clear which interpolation algorithm to choose.}
\LLObservation[1]{The yielded formulas differ in logical strength and length.}
\LLObservation[2]{The weaker the formula, the larger the space covered,
offering the best opportunity to e.g. approximate decision boundaries\Hide{ (cf. Figure~\ref{fig:exp:db})}.}
\LLObservation[3]{Smaller or more focused formulas might be easier to interpret.}
\LLSolution{Use multiple algorithms depending on the current application and flexibly pick suitable outcomes.}

\LLSeparator

\Hide{
\LLProblem{Explanations produced by the \aweaker{} algorithm contain disjunctions.}
\LLObservation[1]{The formulas are conjunctions of disjunctions. \textbf{Reduce} eliminates the whole clauses but cannot simplify them further.}
\LLObservation[2]{Formulas containing disjunctions may be trickier to visualize.}
\LLSolution{Use \aweaker{} if disjunctions are not an issue, and \aweak{} otherwise.}

\LLSeparator
}

\LLProblem{It is not clear how to visualize explanations with high-dimensional impact spaces.}
\LLObservation[1]{Projections (cf. Figure~\ref{fig:exp:itps:merged})
may yield too large area
since the information from the other dimensions is collapsed.}
\LLObservation[2]{Slices fix all other dimensions (cf. Figure~\ref{fig:exp:db}),
but conclusions on those dimensions are very limited.
However,
they
allow direct comparison with decision boundaries or their sampling\Hide{,
in contrast with projections}.}
\LLSolution{Use slices to show local information and to compare with decision boundaries.
Use projections to compare the robustness of explanations.}

\LLSeparator

\Hide{
\LLProblem{What makes one explanation better than another?}
\LLObservation[1]{The larger the space, the more general the explanation.}
\LLObservation[2]{%
The simpler the explanation, the easier it is to communicate.}
\LLSolution{Prefer explanations that fit your current domain-specific use case best.}

\LLSeparator
}

\LLProblem{Sometimes, quantitative and more rigorous comparisons than visualization are needed.}
\LLObservation[1]{Computing the volume of \Hide{complex }high-dimensional spaces is non-trivial.}
\LLObservation[2]{It is easy to check if a space explanation is subsumed by another using the implication of the formulas (in conjunction with~$\psi_\mathcal{D}$).}
\LLObservation[3]{Spaces often intersect but are not entirely subsumed by another.}
\LLSolution{Estimate the quality using multiple metrics.
If needed, include a visual comparison.}

\LLSeparator

\LLProblem[Idea]{What if we compared only selected features?}
\LLObservation[1]{Computing the volume of 2D or 3D spaces is feasible.}
\LLObservation[2]{Quantifier elimination yields exact projections but is expensive.}
\LLObservation[3]{Projections can be approximated via linear programming.}
\LLObservation[4]{Whole explanations can be compared by exhaustively comparing their projections into all particular pairs (or triplets) of features.}
\LLSolution{Comparing only selected features is a simpler problem
and may also enable more thorough comparisons.}

\end{center}

\section{Conclusion and Future Work}
\label{sec:Conclusion}

This paper presented a novel concept of provably correct,
logic-based space explanations of the classification process of neural networks.
The explanations
are associated with complex-shaped spaces
and
capture relations among the features that stem from mathematical proofs,
substantially improving the approximation of decision boundaries
over existing methods.
The Space Explanations concept is supported
by
a flexible framework of algorithms
including efficient Craig interpolation-based techniques and unsatisfiable core extraction to compute an extensive range of different yet meaningful explanations.
On real-world neural networks trained
on practical datasets, we performed a series of experiments and computed explanations of different quality.
The evaluation of our case studies confirms that
our algorithms yield explanations that are more
general than existing explanations.
We shared lessons learned during the extensive experimentation
with the new kind of explanations. 

In future work, we will develop an algorithm to approximate decision boundaries
and to identify reasons for misclassifications
across clusters of the feature space.
We will improve the scalability of our tool \spexplain{}
by using optimizations tailored to NNs,
and aim to handle other NN structures, such as convolutional NNs. 
Finally,
we will apply our method to analyze how decisions evolve across the hidden layers of the network.

\paragraph{Acknowledgements.}
This work was conducted as part of the ``Formal Reasoning on Neural Networks'' project funded by the Hasler Foundation, Switzerland.

\bibliographystyle{splncs04}
\bibliography{references}

\appendix

\section{Appendix}

\subsection{Classifying Neural Networks}
\label{sec:background:nn}
(Classifying) Neural Networks \Hide{(NN) }are a specific class of general classifiers.
Following the notation and definitions,
we assume that the neural networks perform
single-class classification (i.e. each sample point is assigned to exactly one class).
A~neural network is represented
by a~graph $G(E,V)$ with the set of edges~$E$
and the set of vertices $V$, representing the neurons and their weighted connections, respectively.
The neurons are partitioned into layers $V_0,\dots,V_K$
where each~$V_k$ contains neurons
$v_i^{(k)}$, $i \in \{ 1,\dots,n^{(k)} \}$.
Layer~$V_0$ is the \emph{input} layer,
$V_K$ is the \emph{output} layer,
and layers $V_1,\dots,V_{K-1}$ are \emph{hidden} layers.
The edges correspond to the pairs
connecting every neuron of layer~$V_k$
to every neuron in the next layer~$V_{k+1}$,
for all $0 \leq k < K$, i.e., edge~$e_{i,j}^{(k)}$ connects
$i$-th neuron in layer~$k$
with $j$-th neuron in layer~$k+1$.
Every edge is assigned a~numerical weight~$w_{i,j}^{(k)}$.
Moreover,
every neuron~$v_i^{(k)}$ in a~hidden layer is assigned a~bias $b_i^{(k)}$.
Each input neuron~$v_i^{(0)}$
corresponds to feature~$i \in \mathcal{F}$,
represented by feature variable~$x_i \in \mathcal{D}_i$,
and
each output neuron~$v_i^{(K)}$ corresponds to class $c_i$.
Each domain~$\mathcal{D}_i$
is either a~finite set
or a~closed interval $\mathcal{D}_i = [L_i, U_i] $
where both the lower and upper bounds are finite\footnote{
\Hide{%
This is always true as long as the (training) data set is finite.
}%
The restriction to finite intervals is required since a neural network is always trained on a finite set of training data, thus it  learn to expect inputs in a finite range; passing an input (far) outside this range  inevitably cause unpredictable behavior.
}.

\begin{figure}[t!]
    \centering
    \includegraphics[width=0.45\textwidth]{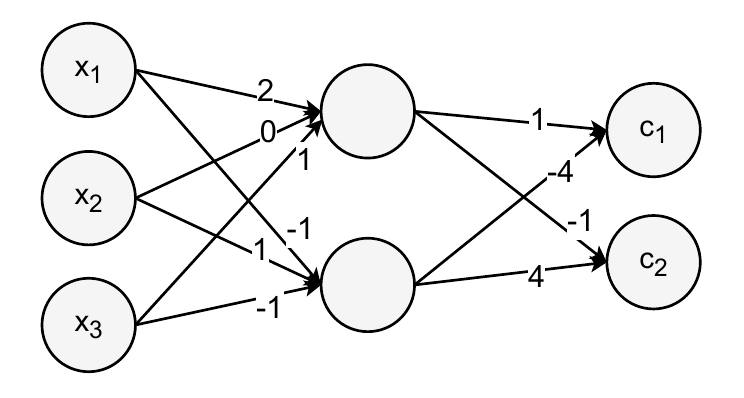}
    \caption{A neural network with one hidden layer}
    \label{fig:nn}
\end{figure}

Given an input sample $\Sample{} = (s_1,\dots,s_m)$,
the activations $x_i^{(k)}$
for each neuron $v_i^{(k)}$ are computed as follows.
In the case of the input layer,
$x_i^{(0)} = s_i$.
In the case of hidden layers,
\begin{equation}
\label{eq:nn:y}
    y_i^{(k)} = \sum_j x_j^{(k-1)}w_{j,i}^{(k-1)} + b_i^{(k)}
\end{equation}
and
$x_i^{(k)} = \Relu(y_i^{(k)})$,
where $\Relu(x) = \max\{x,0\}$ is the rectifier function.
In the case of the output layer,  no activation function is applied\footnote{During the training process, usually the softmax\Hide{~\cite{Rumelhart1986LearningRB}} function is used in the last layer; since this is a monotonic function it can be ignored for the purpose of this paper. 
For details about the \emph{training}, i.e. the parameter estimation of neural networks, we refer to a large number of available resources, e.g.~\cite{Goodfellow_DeepLearning}.}. The class $c_i$ assigned to input sample $\Sample{}$ is given by the output neuron with maximum activation, formally,
\begin{equation}
\label{eq:nn:kappa}
\kappa(\Sample{}) = c_i \iff i = \argmax_{j}\{y_j^{(K)}\}.
\end{equation}


\begin{example}
 \label{example:nn}
Figure \ref{fig:nn} shows a neural network that maps three-dimensional inputs $\mathbf{x}=(x_1, x_2, x_3)$ to two classes $\{c_1,c_2\}$.
The input domains are $\mathcal{D}_i = [0,4]$, therefore, the feature space corresponds to a cube with edges of length 4.
The network consists of 3 input neurons, 2 hidden neurons with $\Relu$ activation, and 2 output neurons. The weights for each edge are labeled accordingly, and for simplicity, biases are set to zero. 
The output is computed as follows:
\begin{alignat*}{4}
x_{1}^{(1)} &= \Relu(2x_1 + x_3)
\qquad
&&o_{c_1} = y_{1}^{(2)} = x_{1}^{(1)} - 4x_{2}^{(1)}
\\
x_2^{(1)} &= \Relu(-x_1 + x_2 -x_3)
\qquad
&&o_{c_2} = y_{2}^{(2)} = -x_{1}^{(1)} + 4x_{2}^{(1)} 
\end{alignat*}
where $o_{c_1}$ and $o_{c_2}$ refer to the values of the output neurons.
If $o_{c_1} \geq o_{c_2}$, then the classification would be $c_1$, otherwise, it would be $c_2$,
e.g., for input sample $\Sample{} = (1,1,3)$, $o_{c_1} = 5$ and $o_{c_2} = -5$ and therefore, $\Sample{}$ is classified as $c_1$.
\end{example}

\subsection{Interpolation-based explanations}
\label{sec:itp-ex}

This section provides additional details on Section~\ref{sec:spaceex}.

\begin{figure}[t!]
    \newcommand*{\subfigurewidthratio}{0.32}
    \newcommand*{\subfiguregraphicswidthratio}{1.0}
    \newcommand*{\trimleft}{1cm}
    \newcommand*{\trimright}{\trimleft}
    \newcommand*{\trimbottom}{2cm}
    \newcommand*{\trimtop}{\trimbottom}

    \begin{center}
        \begin{subfigure}{\subfigurewidthratio\textwidth}
            \includegraphics[trim={\trimleft{} \trimbottom{} \trimright{} \trimtop{}},clip,width=\subfiguregraphicswidthratio\columnwidth]{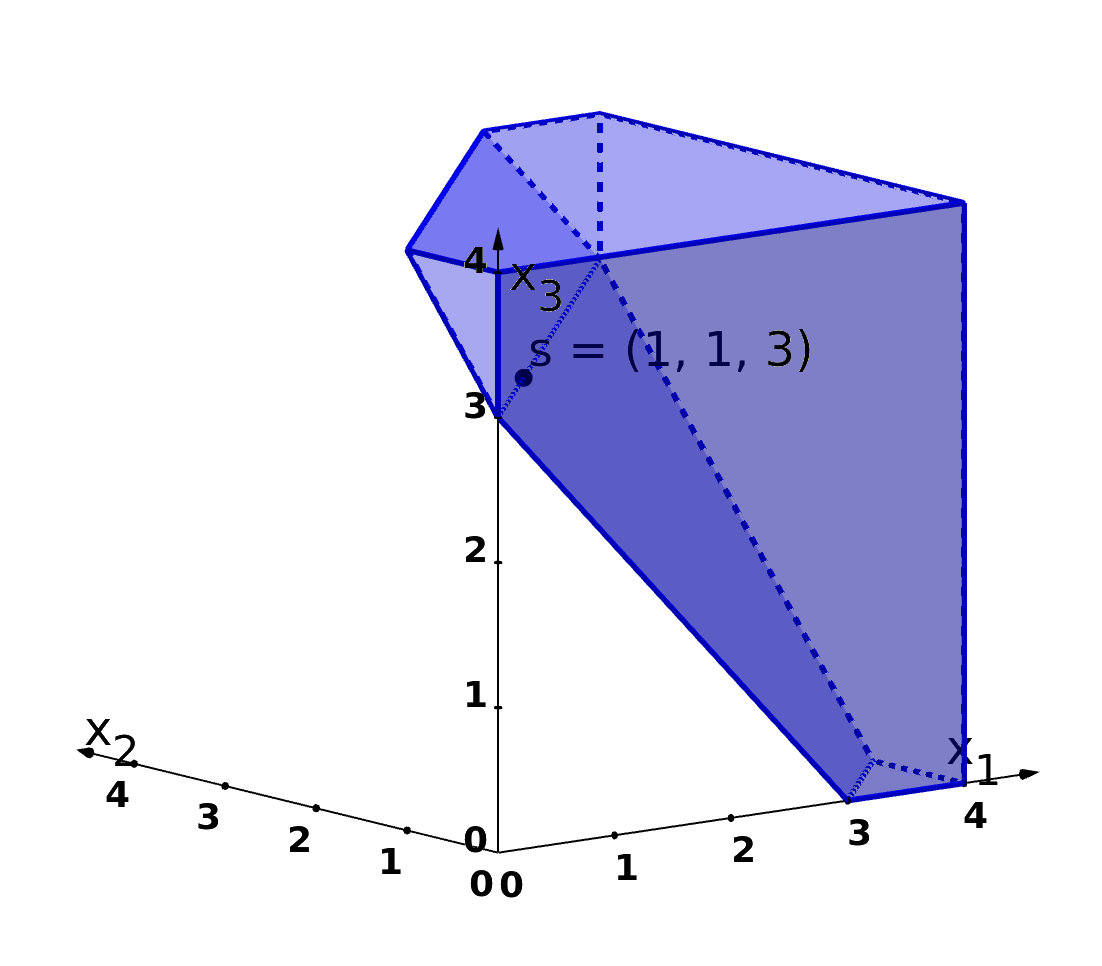}
            \caption{$\varphi_1 = x_1 - x_2 + x_3 \geq 3$}
            \label{fig:spaceex:1}
        \end{subfigure}
        \hfill
        \begin{subfigure}{\subfigurewidthratio\textwidth}
            \includegraphics[trim={\trimleft{} \trimbottom{} \trimright{} \trimtop{}},clip,width=\subfiguregraphicswidthratio\columnwidth]{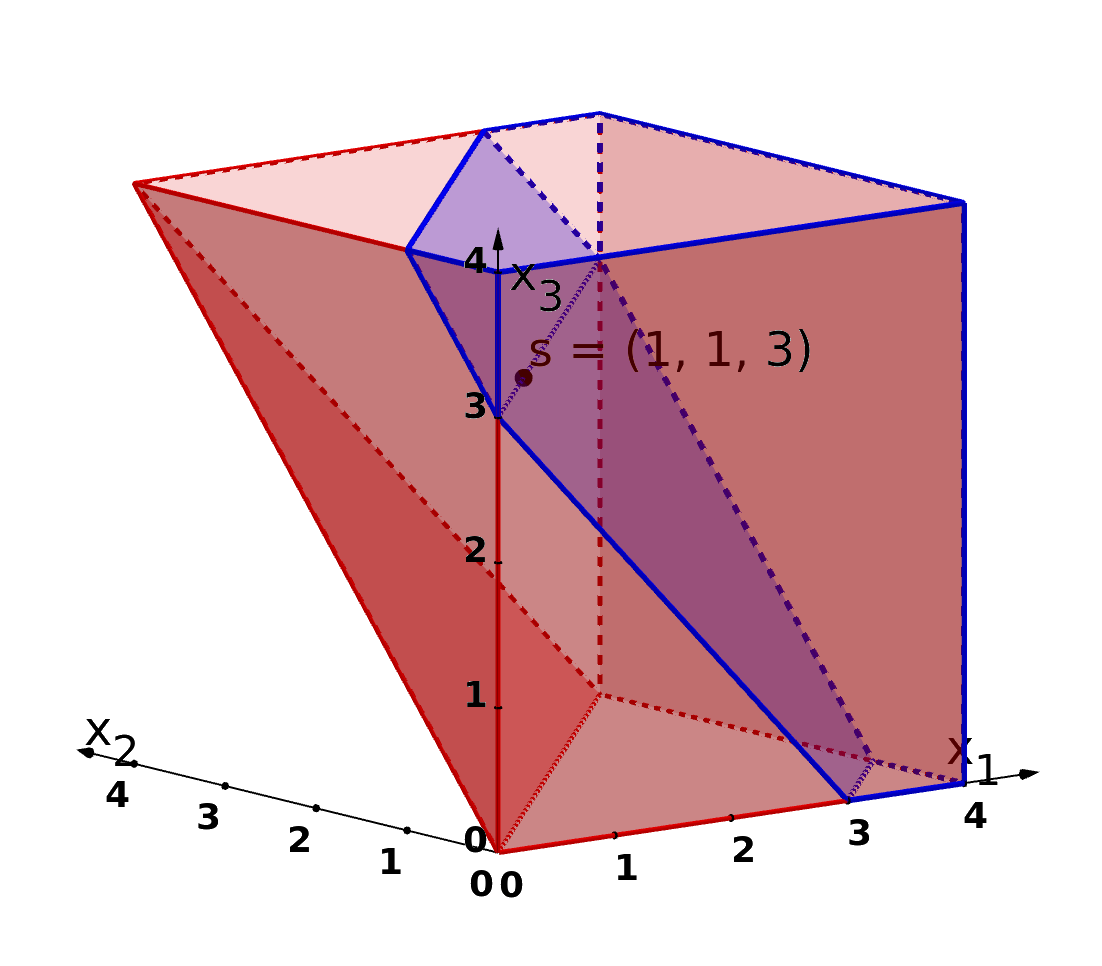}
            \caption{$\varphi_1^* = x_1 - x_2 + x_3 > 0$}
            \label{fig:spaceex:2}
        \end{subfigure}
        \hfill
        \begin{subfigure}{\subfigurewidthratio\textwidth}
            \includegraphics[trim={\trimleft{} \trimbottom{} \trimright{} \trimtop{}},clip,width=\subfiguregraphicswidthratio\columnwidth]{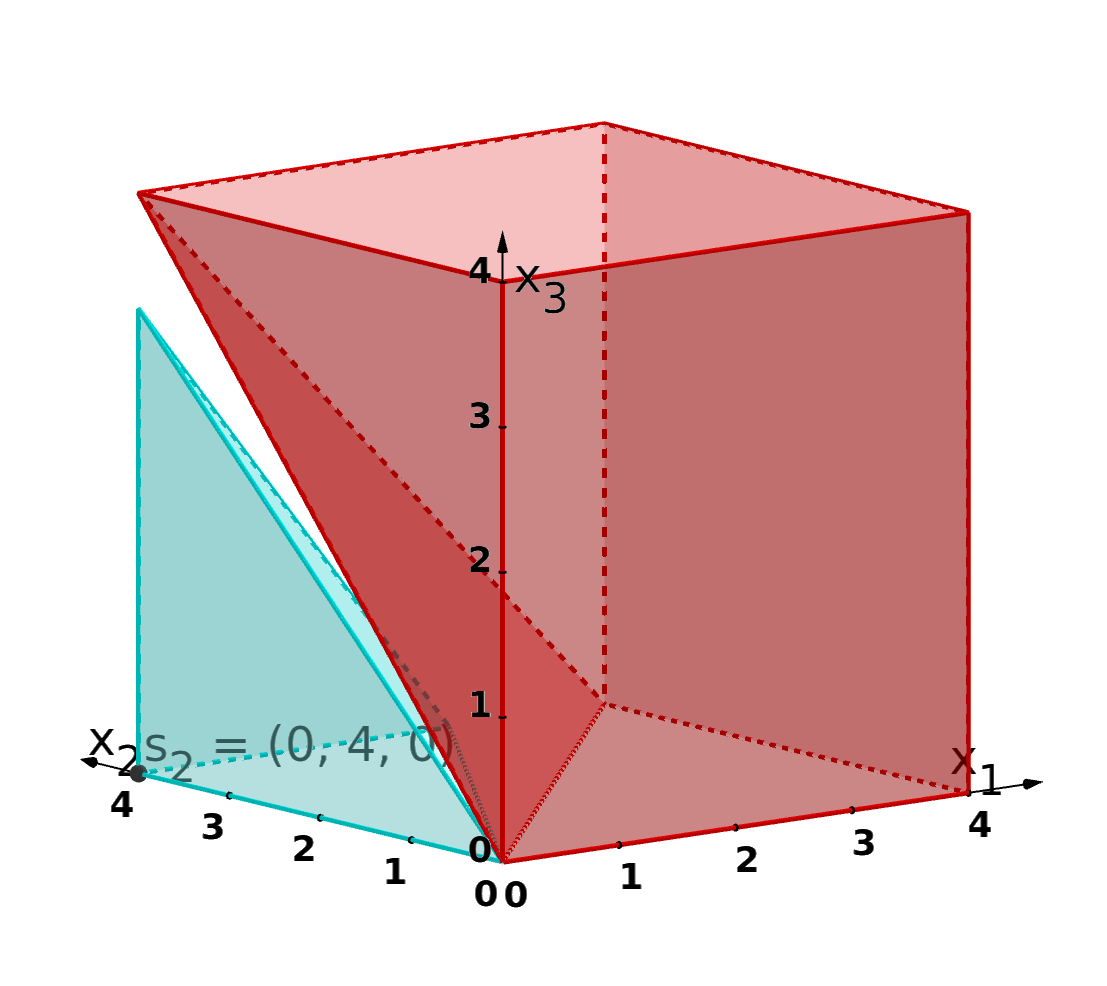}
            \caption{$\varphi_2 = x_1 - \frac{2}{3} x_2 + \frac{5}{6} x_3 < \frac{1}{384}$}
            \label{fig:spaceex:3}
        \end{subfigure}
    \end{center}

    \caption{Impact spaces associated with three interpolation-based explanations
    for the network in Example~\ref{example:nn}
    with
    the feature space $\mathbb{F} = [0,4]^3$.
    Figure~\subref*{fig:spaceex:1}
    represents explanation~$\varphi_1$ (blue)
    for the sample point $\Sample{} = (1,1,3)$,
    and
    Figure~\subref*{fig:spaceex:2}
    represents a~weaker explanation~$\varphi_1^*$ (red).
    Figure~\subref*{fig:spaceex:3}
    shows
    explanation~$\varphi_2$ (cyan)
    for a~different sample point $\Sample{}_2 = (0,4,0)$
    classified into class~$c_2$.
    The area between the impact spaces~$\ImpSpace{\varphi_1^*}$
    and~$\ImpSpace{\varphi_2}$
    contains the decision boundary.
    }
    \label{fig:spaceex}
\end{figure}

\begin{example}
    \label{example:spaceex:itp}
    Given the toy neural network in Example~\ref{example:nn},
    we depict three automatically computed interpolation-based explanations
    in Figure~\ref{fig:spaceex}.
    Explanation~$\varphi_1$
    of the sample point $\Sample{} = (1,1,3)$
    classified to class~$c_1$
    captures a~mutual relationship among all features.
    The formula resembles the term~$x_2^{(1)}$ in Example~\ref{example:nn}\Hide{,
    confirming the claims in item~\ref{it:itp:proof} in Section~\ref{sec:spaceex}
    (and in item~\ref{it:spaceex:relship})}.

    Next,
    we obtained explanation~$\varphi_1^*$
    using the same sample point~$\SampleExpl$
    but a~different (logically weaker) interpolation algorithm
    than with~$\varphi_1$\Hide{,
    addressing item~\ref{it:itp:flex}}.
    Moreover,
    $\varphi_1^*$ can also be computed on top of~$\varphi_1$
    instead of~$\SampleExpl$\Hide{,
    addressing item~\ref{it:itp:general}}.

    Finally,
    we obtained
    explanation~$\varphi_2$
    similarly as~$\varphi_1^*$
    but using a~different sample point $\Sample{}_2 = (0,4,0)$
    classified to class~$c_2$.
    Since almost the whole feature space is already covered,
    the impact spaces
    are close \emph{underapproximations} of both \emph{class spaces}.
    Consequently,
    the void space
    between~$\ImpSpace{\varphi_1}$ and $\ImpSpace{\varphi_2}$
    is a~guaranteed \emph{overapproximation} of the \emph{decision boundary}\Hide{,
    addressing item~\ref{it:spaceex:shape} in Section~\ref{sec:spaceex}}.
\end{example}

\begin{figure}
    \centering
    \includegraphics[trim={1cm 2.5cm 1cm 3.5cm},clip,width=0.3\textwidth]{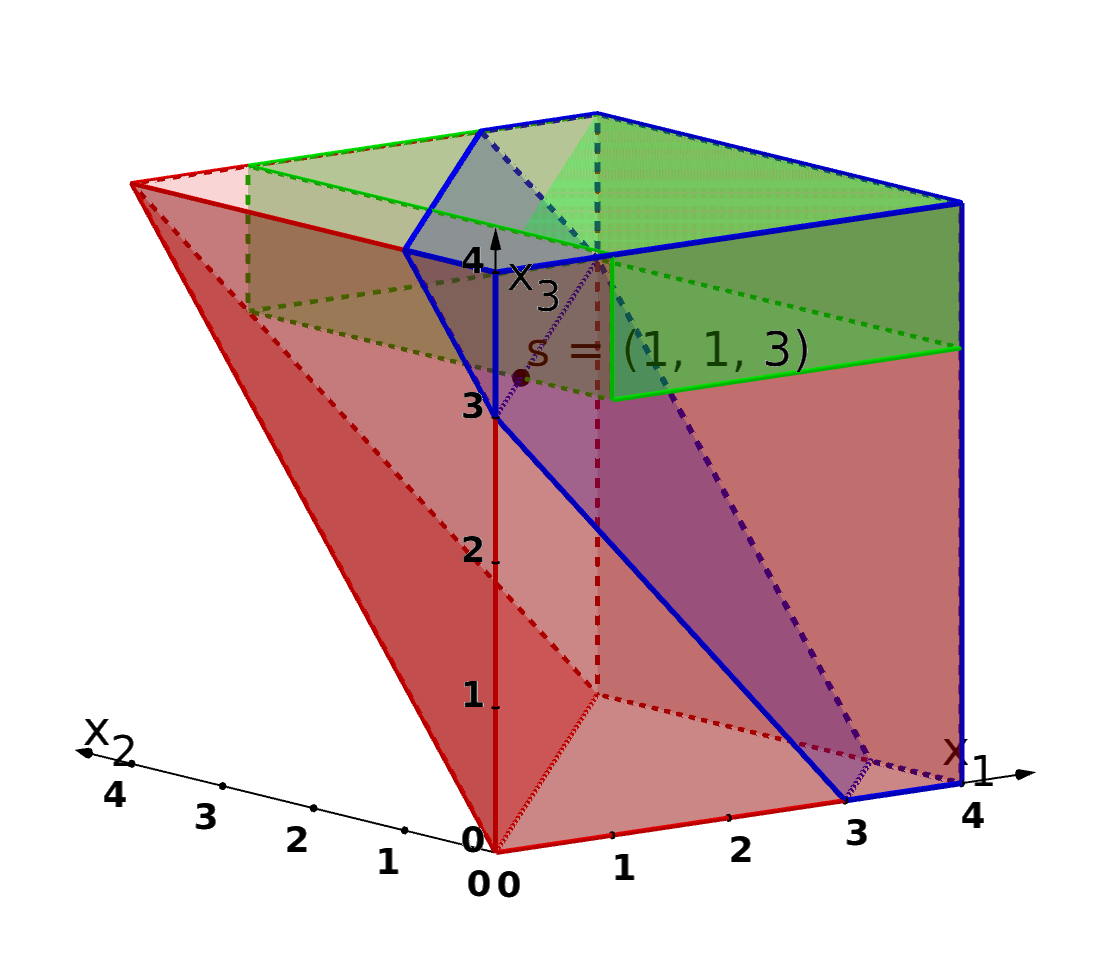}
    \caption{Comparison to interval explanation (green)}
    \label{fig:spaceex:inflated}
\end{figure}

In Figure~\ref{fig:spaceex:inflated},
we compare our explanations from Figure~\ref{fig:spaceex:2}
with a~state-of-the-art explanation~\cite{Ignatiev:24}
that cannot be expanded further feature-wise,
using ordering of the features $(x_2, x_1, x_3)$.
Not only the space is much smaller,
but the explanation also provides much less information
with no feature relationships.

\end{document}